\documentclass[a4paper,twoside]{article}

\usepackage{epsfig}
\usepackage{subcaption}
\usepackage{calc}
\usepackage{amssymb}
\usepackage{amstext}
\usepackage{amsmath}
\usepackage{amsthm}
\usepackage{multicol}
\usepackage{pslatex}
\usepackage{apalike}
\usepackage[bottom]{footmisc}

\usepackage{cite}
\usepackage{booktabs}
\usepackage{xspace}

\usepackage{SCITEPRESS}     % Please add other packages that you may need BEFORE the SCITEPRESS.sty package.

\def\etal{\emph{et al.}\xspace}
\def\ie{\emph{ie.}\xspace}

\newcommand{\hide}[1]{}
\newcommand{\tabincell}[2]{\begin{tabular}{@{}#1@{}}#2\end{tabular}}

\begin{document}

\title{Rethinking the Backbone Architecture for Tiny Object Detection}

\author{\authorname{Jinlai Ning\sup{1}\orcidAuthor{0000-0002-0460-7657}, Haoyan Guan\sup{1}\orcidAuthor{0000-0003-1936-2442} and Michael Spratling \sup{1}\orcidAuthor{0000-0001-9531-2813}}
\affiliation{\sup{1}Department of Informatics, King's College London, London, UK}
\email{\{jinlai.ning\, haoyan.guan\, michael.spratling\}@kcl.ac.uk}
}


\abstract{Tiny object detection has become an active area of research because images with tiny targets are common in several important real-world scenarios. However, existing tiny object detection methods use standard deep neural networks as their backbone architecture. We argue that such backbones are inappropriate for detecting tiny objects as they are designed for the classification of larger objects, and do not have the spatial resolution to identify small targets. Specifically, such backbones use max-pooling or a large stride at early stages in the architecture. This produces lower resolution feature-maps that can be efficiently processed by subsequent layers. However, such low-resolution feature-maps do not contain information that can reliably discriminate tiny objects. To solve this problem we design ``bottom-heavy'' versions of backbones that allocate more resources to processing higher-resolution features without introducing any additional computational burden overall. We also investigate if pre-training these backbones on images of appropriate size, using CIFAR100 and ImageNet32, can further improve performance on tiny object detection. Results on TinyPerson and WiderFace show that detectors with our proposed backbones achieve better results than the current state-of-the-art methods.}

\onecolumn \maketitle \normalsize \setcounter{footnote}{0} \vfill

\section{\uppercase{Introduction}}
\label{sec:introduction}

% introduce tiny object detection
Tiny object detection is a sub-field of object detection in the field of computer vision, and has many applications including maritime search-and-rescue, surveillance, and driving assistance. It is an active area of research because classical methods of standard object detection \cite{fasterrcnn, ssd, retinanet, cascade, yolov3, fcos} that work well on datasets such as MS COCO \cite{mscoco} and Pascal VOC \cite{pascalvoc} perform poorly on tiny object datasets such as TinyPerson \cite{tinyperson} and WiderFace \cite{widerface}. Detecting tiny objects is still a challenging task for these standard methods. As a result, many methods have been designed specifically for tiny object detection \cite{survey}. As described in \cite{survey}, these existing methods adapt standard object detection frameworks to be more suitable for tiny object detection by: using super-resolution techniques \cite{prior_info_hr, pyramidbox}; exploiting contextual information \cite{sspnet}; using data augmentation techniques \cite{cutmix, tinyperson, sm+}; employing multi-scale representation learning approaches \cite{fpn, fusionfactor}; using anchor mechanisms more appropriate for small objects \cite{s3fd, lsfhi}; designing training strategies specific to small objects \cite{sniper, isod}; or using loss functions specific for small and tiny objects \cite{feedback, focalloss}. In this article we propose a different approach that does not fit into any of these categories.

\begin{figure*}[t]
\centering
\subfloat[ResNet50]{\includegraphics[scale=0.45]{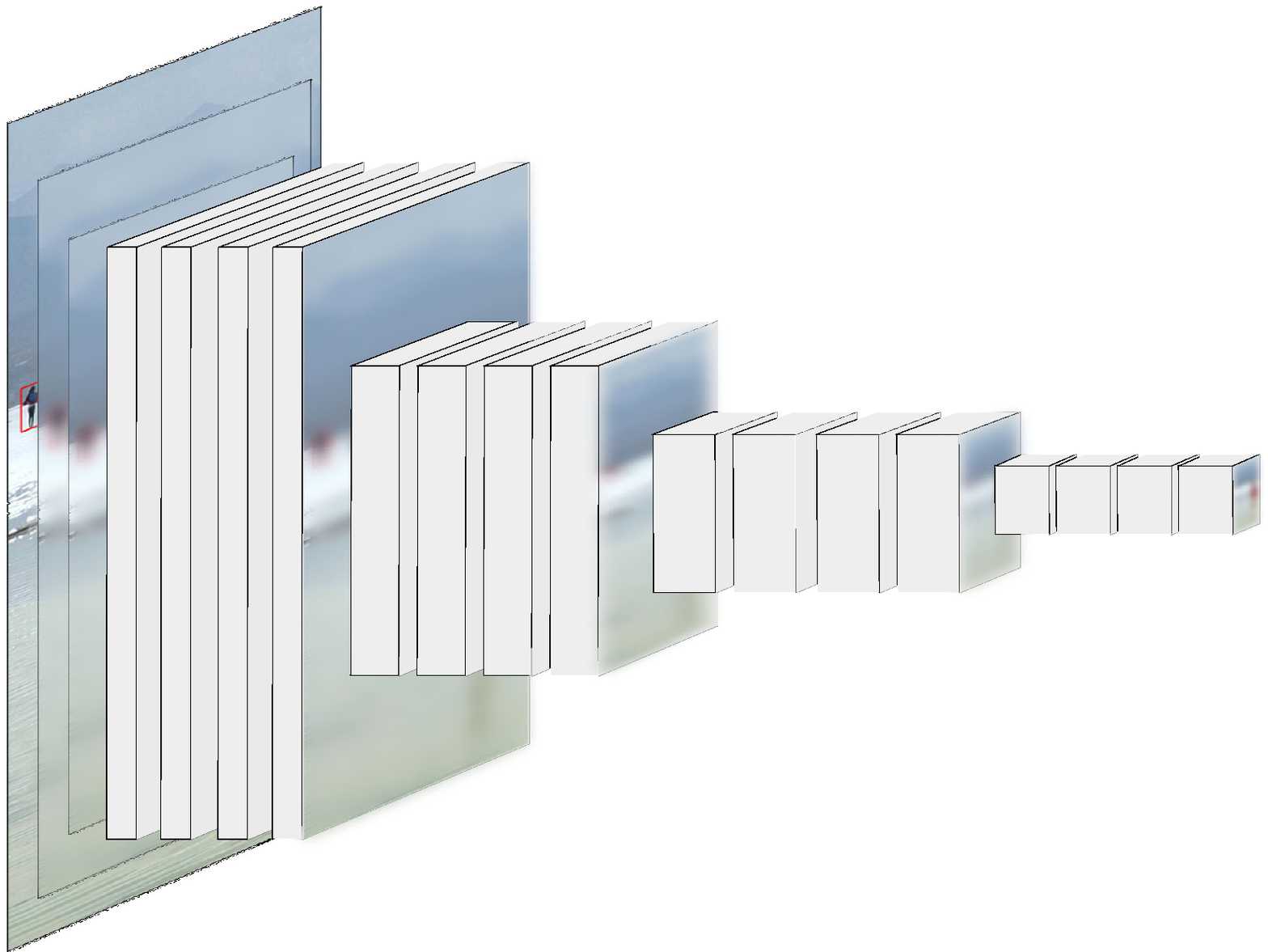}}
\subfloat[BH-ResNet50]{\includegraphics[scale=0.45]{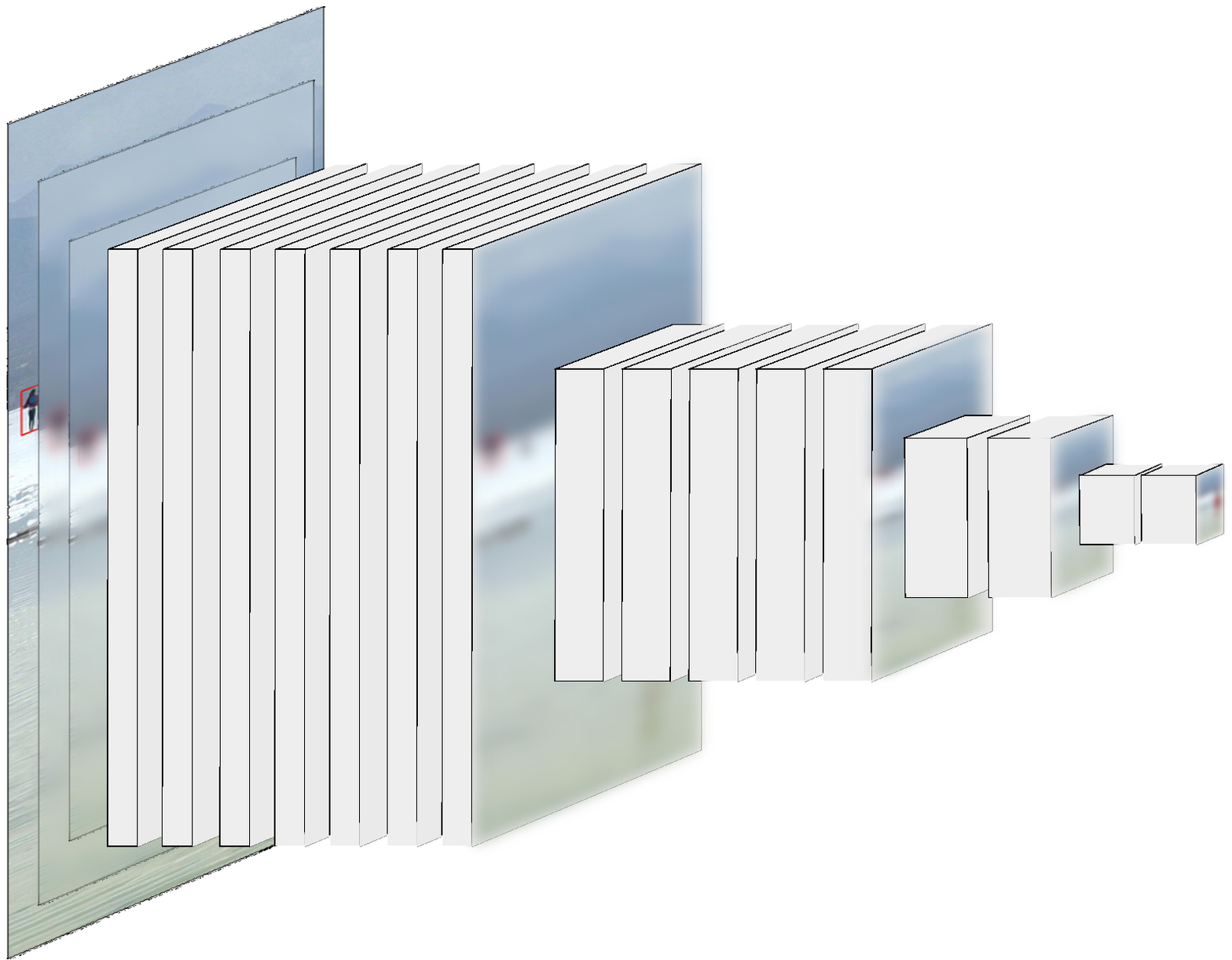}}
\caption{An illustration of (a) an existing backbone, and (b) our modified backbone. Our modified backbone delays down-sampling and changes the number of layers at different depths. 
}
\label{fig:arch}
\end{figure*}

Although previous approaches vary from each other in specific details, they all depend on standard deep neural network backbones, such as ResNet \cite{resnet}, to extract features. The features are then used for feature fusion, position regression and category classification. Despite the impressive results previous algorithms have obtained, we believe that by re-deploying the same standard backbones as are used for general object detection \cite{fasterrcnn, retinanet, cascade, fcos}, these methods suffer from poor feature extraction for tiny object detection. The standard backbone architectures down-sample the feature maps rapidly in the first few layers, and this down-sampling removes much of the information about tiny objects that were present in the original image. As a result, subsequent layers in the backbone, and all the methods for improving tiny object detection that fall into the seven categories mentioned in the preceding paragraph, can only work with features that are relatively uninformative and poor at identifying tiny objects. 

% introduce why down-sampling matters
Down-sampling is a widely applied operation that has proven highly effective in convolutional neural networks (CNNs). It can be achieved using pooling layers that summarize the features in each patch, or by using a convolutional layer with a stride greater than one \cite{sfs}. Down-sampling can improve  translation invariance, avoid over-fitting, and decrease computational costs. For tiny object detection, down-sampling itself is not an issue. Rather it is the improper use of down-sampling that results in poor feature-extraction.
%introduce the side effect of down-sampling, especially on tiny objects 
Tiny objects occupy very few pixels and down-sampling could potentially remove important features that identify such objects. The only way to preserve information about small features is for convolutional filters in the earliest layers to encode these features and pass this information on to subsequent layers. However, in existing backbones the number of convolutional filters in the early layers is kept to a minimum to reduce the computational burden, and this likely means that not all discriminative features of tiny objects are able to be preserved. For instance, ResNets decrease feature map size by a factor of 4 in less than or equal to 2 layers of convolutions. Using such backbones to handle tiny object detection is likely to result in information about tiny objects disappearing in the feature maps before it is fully extracted. 

% introduce previous research abut information loss 
Outside the domain of object detection, previous work has addressed information loss caused by down-sampling in a number of different ways. Zhao \etal \cite{rdmshift} introduced random shifting in down-sampling during the training process to suppress contextual information loss without extra calculation cost. Ren \etal \cite{lossrj} proposed a low-rank feature recovery module to try to recover the lost information. In other sub-fields of computer vision, such as hyper-spectral target detection \cite{lossab}, and pluralistic image in-painting \cite{losslq}, several methods have been designed to handle information loss. Similarly, the chess-playing AI, AlphaGo \cite{alphago}, did not use layers with strides bigger than one to avoid losing any spatial information about the chessboard. However, in the domain of tiny object detection, this problem has previously not been considered or has been overlooked due to the convenience of using the same backbone as has been used in general-purpose object detectors \cite{fasterrcnn, ssd, yolo}.

\begin{table*}[tb]
\centering
\caption{The size of objects in several typical general object detection datasets (above the double lines) and tiny object detection datasets (below the double lines). Absolute size is defined as the square root of the object’s absolute bounding box area, measured in pixels. It is reported as a mean$\pm$standard deviation.}
\label{tab:dataset}
\begin{tabular}{l|l|l}
\toprule[1.5pt]
\textbf{Dataset}&\textbf{Absolute size}&\textbf{Proportion of small/tiny objects}\\ \hline
MS COCO&$99.5\pm107.5$&   32\% (with absolute size$\leq32$ pixels)\\ %\hline
PASCAL VOC&---&           10\% (with absolute size$\leq32$ pixels)\\ %\hline 
CityPersons&$79.8\pm67.5$& 8\% (with absolute size$\leq20$ pixels)\\ \hline \hline
TinyPerson&$18.0\pm17.4$& 73\% (with absolute size$\leq20$ pixels)\\ %\hline
WiderFace&$32.8\pm52.7$&  56\% (with absolute size$\leq20$ pixels)\\ 
\bottomrule[1.5pt]
\end{tabular}%
\end{table*}

% what we have done
In this paper we address this neglected problem and propose alternative backbone architectures that are more appropriate for tiny object detection. We show that simple changes to the architecture (as illustrated in Fig.~\ref{fig:arch}) that delay down-sampling operations, and move more convolutional filters to earlier layers where they can process high-resolution information, can bring clear improvement to object detection performance. We demonstrate the effectiveness of this strategy by making the same modifications to two different backbone architectures: ResNet \cite{resnet} and HRNet \cite{hrnet}. Experiments on TinyPerson \cite{tinyperson} and WiderFace \cite{widerface} demonstrate that replacing standard backbones with our modified versions always results in an improvement in performance for a number of different object detection frameworks. This is achieved despite the modified backbones having fewer parameters than the corresponding original ones. These results support our claim that unsuitable use of down-sampling occurs in the backbones commonly used for tiny object detection and that this problem should be taken seriously to help improve detection accuracy. 

\begin{table*}[tb]
\centering
\caption{Detection performance of previous methods, including the current state-of-the-art methods, on TinyPerson, sorted by performance on the $mAP_{tiny}^{50}$ metric.}
\label{tab:sota_tp}
%\resizebox{\textwidth}{!}{%
\begin{tabular}{l|l|l|l|l|l}
\toprule[1.5pt]
\textbf{Method}&\textbf{$mAP_{tiny}^{50}$}&\textbf{$mAP_{tiny1}^{50}$}&\textbf{$mAP_{tiny2}^{50}$}&\textbf{$mAP_{tiny3}^{50}$}&\textbf{$mAP_{small}^{50}$} \\ \hline
Faster R-CNN-FPN \cite{fasterrcnn}&47.35&30.25&51.58&58.95&63.18\\ 
Faster R-CNN-FPN with S-$\alpha$ \cite{fusionfactor}&48.39&31.68&52.20&60.01&65.15\\
Faster R-CNN-RFLA \cite{rfla}&48.86&30.35&54.15&61.28&66.69\\
Faster R-CNN-FPN-SM \cite{tinyperson}&51.33&33.91&55.16&62.58&66.96\\
Faster R-CNN-FPN-SM+ \cite{sm+}&51.46&33.74&55.32&62.95&67.37\\
RetinaNet-SSPNet \cite{sspnet}&54.66&42.72&60.16&61.52&65.24\\
Faster R-CNN with SFRF \cite{sfrf}&57.24&51.59&64.51&67.78&65.33\\ 
Cascade R-CNN-SSPNet \cite{sspnet}&58.59&45.75&62.03&65.83&71.80\\
Faster R-CNN-SSPNet \cite{sspnet}&59.13&47.56&62.36&66.15&71.17\\ 
\bottomrule[1.5pt]
\end{tabular}%
%}
\end{table*}

%\resizebox{0.65\textwidth}{!}{%
\begin{table*}[tb]
\centering
\caption{Detection performance of previous methods, including the current state-of-the-art methods, on WiderFace, sorted by performance on the $mAP_{medium}$ metric.}
\label{tab:sota_wf}
\begin{tabular}{l|p{1.5cm}|p{1.5cm}|p{1.5cm}}
\toprule[1.5pt]
\textbf{Method}&$mAP_{easy}$&$mAP_{medium}$&$mAP_{hard}$ \\ \hline
HR \cite{hr}&92.5&91.0&80.6\\
S3FD \cite{s3fd}&93.7&92.4&85.2\\
FaceGAN \cite{face_gan}&94.4&93.3&87.3\\
SFA \cite{sfa}&94.9&93.6&86.6\\
LSFHI \cite{lsfhi}&95.7&94.9&89.7\\
Pyramid-Box \cite{pyramidbox}&96.1&95.0&88.9\\
RetinaFace \cite{retinaface}&96.5&95.6&90.4\\
TinaFace \cite{tinaface}&96.3&95.7&93.1\\
\bottomrule[1.5pt]
\end{tabular}%
%}
\end{table*}

% \vfill
\section{\uppercase{Related Work}}

% object detection
Object detection aims to locate, using a bounding box, and predict the category of each object in an image. The majority of state-of-the-art algorithms are based on deep learning techniques \cite{fasterrcnn,effdet,fcos,syolo4,swintra}, although classic image processing methods have also made contributions in the early years \cite{hogdet,dpm}. CNN-based networks have dominated the object detection field for many years. These methods can be classified as one-stage
\cite{yolo1,yolo4,ssd,fcos} or two-stage \cite{rcnn, sppnet, fasterrcnn} algorithms, depending on whether or not detection is performed by one end-to-end network. Rather than designing bespoke backbone architectures, methods in both groups use a backbone that is a standard CNN architecture that has been pre-trained on a classification task, and then had the fully-connected layers removed. 

% Datasets for object detection and tiny object detection
Large datasets are the basic resources required for deep learning-based methods. For object detection, MS COCO \cite{mscoco} and PASCAL VOC \cite{pascalvoc} are widely used as benchmarks to evaluate the performance of different algorithms. Both of them contain a variety of natural images, containing a large number of different categories, and can be used to test how object-detection methods perform on general tasks. 

% give a brief description of tod
Tiny object detection is a sub-field of object detection. It focuses on detecting tiny objects in images because such objects are common in real-world scenarios, but are hard to detect with standard object detection methods. The main difference between tiny object detection datasets such as TinyPerson \cite{tinyperson} and WiderFace \cite{widerface} and general object detection datasets is the scale of the target objects. As shown in Table~\ref{tab:dataset}, tiny object datasets contain more small and tiny objects while general datasets contain objects at a wider range of scales. 
% Previous methods for tiny detection
As mentioned in the Introduction, previous methods for tiny object detection fall into seven categories \cite{survey}. Representative examples from each category will be reviewed in the following paragraphs, with results summarized in Tables~\ref{tab:sota_tp} and~\ref{tab:sota_wf}.

Super-resolution techniques increase the resolution of the image, to enable standard techniques for larger object detection to be applied successfully to images containing tiny targets. Increasing the resolution can be done for the whole image, for example using a generative neural network \cite{face_gan}, or can be used on a small region of interests \cite{prior_info_hr}.

Contextual information (\ie features from the surrounding region of the image) can be used to help detect small objects. Hence, many techniques include context in the computation. One example is the scale selection pyramid network (SSPNet) that consists of a context attention module (CAM), scale enhancement module (SEM), and scale selection module (SSM) \cite{sspnet}. CAM includes context information by generating hierarchical attention heat-maps. SEM ensures feature maps of different scales only focus on objects of suitable scale rather than other objects and the background. SSM exploits feature fusion between shallow and deep features to keep the information shared by different layers. 

Data augmentation aims to improve performance by extending the size of training dataset through identity-preserving image transformations. Yu \etal \cite{tinyperson} proposed a simple yet efficient augmentation approach named scale match. This method can transform the distribution of object sizes from a general dataset to be similar to a task-specific dataset. In that paper, MS COCO was transformed with scale match and then used as additional data for training a detector for the TinyPerson detection task. One step in scale match is to sample a size-bin from the histogram of object sizes, this sampling is done at random with the probability based on a given image in the external dataset. To avoid the range of sampled size-bins having a big difference with the size of objects in the external image, monotone scale match (MSM) is utilized to sample the range of sizes from small to big monotonically. Jiang \etal \cite{sm+} proposed an enhanced version of scale match, SM+, that improves the scale match from image level to instance level and uses probabilistic structure inpainting (PSI) to handle the background.

Multi-scale representation learning aims to enhance detection performance by making use of features at different scales, represented in different layers of the CNN. This technique is also commonly applied in general object detection, for example through the feature pyramid network (FPN) \cite{fpn}. Gong \etal \cite{fusionfactor} argue that the FPN brings not only a positive impact but also a negative one caused by the top-down connections. They design a statistic-based fusion factor to adaptively adjust the weight of different layers during the feature fusion. Tang \etal \cite{pyramidbox} argue that not all high-level semantic features are really helpful to smaller targets. They therefore modify FPN to a low-level feature pyramid network (LFPN) that starts the top-down structure from a middle layer rather than the high layer. Liu \etal \cite{sfrf} devise a feature rescaling and fusion (SFRF) network that selects and generates a new resized feature map with a high-density distribution of tiny objects through the use of a Nonparametric Adaptive Dense Perceiving Algorithm (NADPA) module.

Anchors are predefined locations where objects might be detected and are widely applied in two-step detectors such as Faster R-CNN \cite{fasterrcnn} and Cascade R-CNN \cite{cascade}. Zhang \etal \cite{s3fd} guarantees that different scales of anchor have the same density on the image, so that various scales can approximately match the same number of anchors. They also propose a scale compensation anchor matching strategy to ensure that all scales have enough anchors matched. Deng \etal \cite{retinaface} specify different sizes of anchors in different layers of features to ensure all sizes of objects could be matched properly. Zhang \etal \cite{lsfhi} develop a single-level face detection framework to specifically detect small faces, which uses different dilation rates for anchors with different sizes and performed anchor-level hard example mining. Xu \etal \cite{rfla} design a novel label assignment method to mitigate the problems of lack of positive samples and gap between the uniformly distributed prior anchors and the Gaussian distributed receptive field.

Adapting the training strategy to be more appropriate for tiny object detection can be beneficial. For example, Luo \etal \cite{sfa} propose a multi-branch small face attention (SFA) detector. It trains the model with multi-scale training images to improve robustness and performance. Hu \etal \cite{hr} train separate face detectors efficiently by utilizing features extracted from multiple layers of a single feature hierarchy for different scales.

\begin{figure}[t]
\centering
\subfloat[ResNet50]{\includegraphics[scale=0.45]{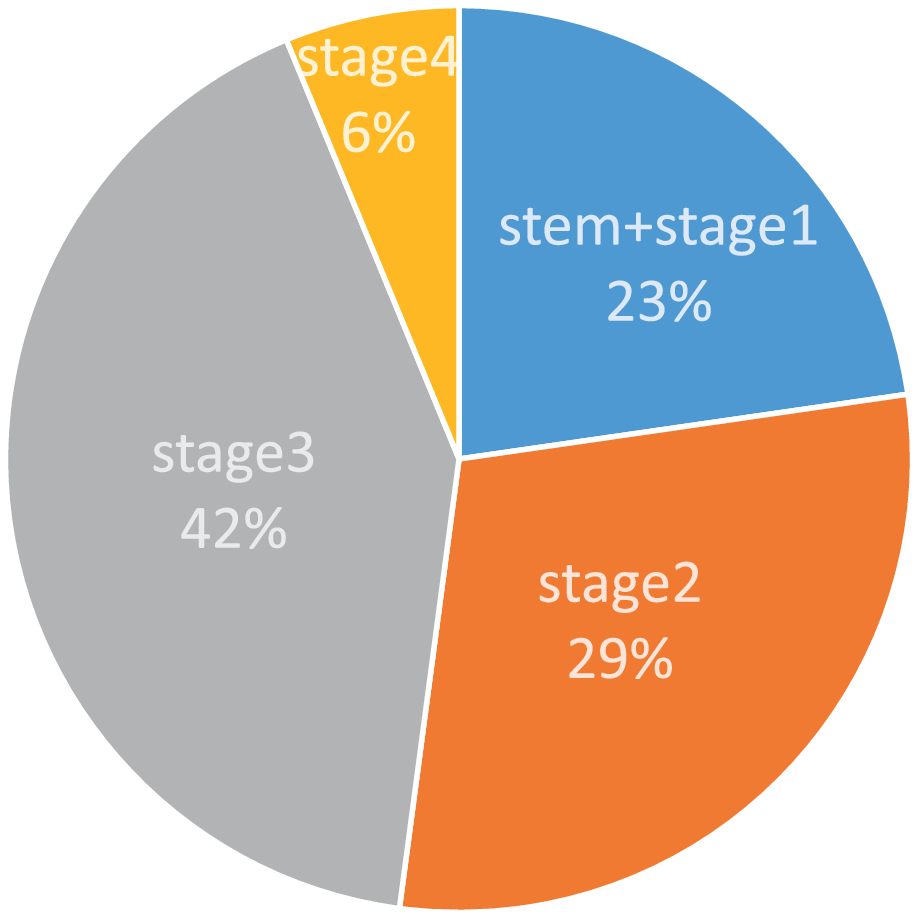}}\\
\subfloat[BH-ResNet50]{\includegraphics[scale=0.45]{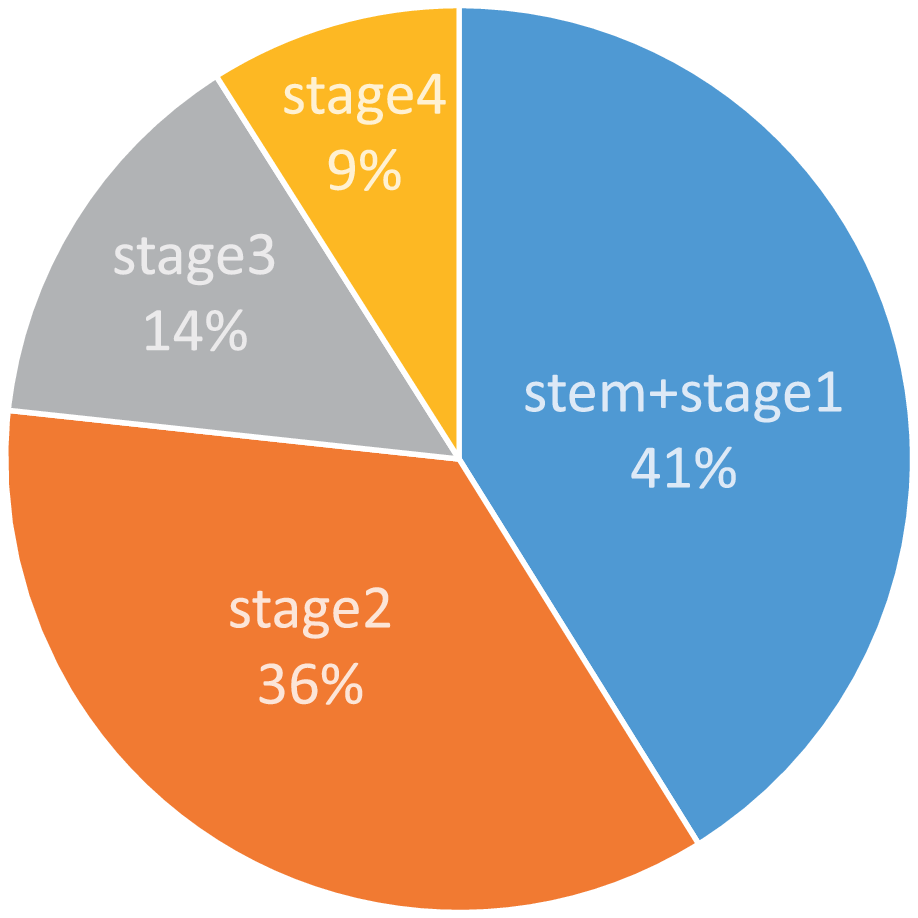}}
\caption{The proportion of FLOPs per stage for (a) standard ResNet50, and (b) our modified BH-ResNet50.}
\label{fig:proportion}
\end{figure}

\begin{table*}[tb]
\centering
\caption{The architectures of three standard deep neural network backbones (ResNet50, HRNet32, and HRNet18) and the proposed bottom-heavy (BH) versions. Convolutional layers are specified using the notation width $\times$ height, number of channels, and stride. Pooling layers are specified using the notation width $\times$ height, type, and stride, where type is M for max pooling. Repeated blocks are shown using squared brackets with the number of repeats indicated by the number following the  multiply sign after the brackets. Where stride is unspecified it is equal to one. '*' indicates the stride of that layer equals two but only for the first block in a set of repeated layers. Rather than illustrating all parallel branches of HRNet, only the deepest branch is shown for clarity. The number of giga-floating-point operations (GFLOPs) is calculated for an input size of $640$ pixels $\times512$ pixels.}
\label{tab:shallow}
\resizebox{\textwidth}{!}{%
\tabcolsep2pt
\begin{tabular}{p{14mm}|c|c|c|c|c|r}
\toprule[1.5pt]
&\textbf{Stem}&\textbf{Stage 1}&\textbf{Stage 2}&\textbf{Stage 3}&\textbf{Stage 4}&\tabincell{l}{\textbf{Params} \&\\\textbf{GFLOPs}}\\ 
%&&&&&&(M)&\\
\hline
ResNet50
&\tabincell{l}{$7\times7,64,2$\\$3\times3,M,2$}
&\tabincell{c}{$\left[\begin{array}{cc}1\times1,&64\\3\times3,&64\\1\times1,&256\end{array}\right]\times3$}
&\tabincell{c}{$\left[\begin{array}{cc}1\times1,&128\\3\times3,&128,2^{*}\\1\times1,&512\end{array}\right]\times4$}&\tabincell{c}{$\left[\begin{array}{cc}1\times1,&256\\3\times3,&256,2^{*}\\1\times1,&1024\end{array}\right]\times6$}&\tabincell{c}{$\left[\begin{array}{cc}1\times1,&512\\3\times3,&512^{*}\\1\times1,&2048\end{array}\right]\times3$}&\tabincell{l}{23.23M\\26.91}\\ \hline
\tabincell{l}{BH-\\ResNet50} &\tabincell{l}{$3\times3,64,1$\\$3\times3,M,2$}&\tabincell{c}{$\left[\begin{array}{cc}1\times1,&64\\3\times3,&64,2^{*}\\1\times1,&256\end{array}\right]\times7$}&\tabincell{c}{$\left[\begin{array}{cc}1\times1,&128\\3\times3,&128,2^{*}\\1\times1,&512\end{array}\right]\times6$}&\tabincell{c}{$\left[\begin{array}{cc}1\times1,&256\\3\times3,&256,2^{*}\\1\times1,&1024\end{array}\right]\times2$}&\tabincell{c}{$\left[\begin{array}{cc}1\times1,&512\\3\times3,&512,2^{*}\\1\times1,&2048\end{array}\right]\times1$}&\tabincell{l}{10.91M\\27.12}\\ \hline \hline 
HRNet32&\tabincell{l}{$3\times3,64,2$\\$3\times3,64,2$}&\tabincell{c}{$\left[\begin{array}{cc}3\times3,&64\\3\times3,&64\end{array}\right]\times4$}&\tabincell{c}{$\left[\begin{array}{cc}3\times3,&64\\3\times3,&64\end{array}\right]\times4$}&\tabincell{c}{$\left[\begin{array}{cc}3\times3,&128\\3\times3,&128\end{array}\right]\times16$}&\tabincell{c}{$\left[\begin{array}{cc}3\times3,&256\\3\times3,&256\end{array}\right]\times12$}&\tabincell{l}{29.31M\\51.93}\\  \hline 
\tabincell{l}{BH-\\HRNet32} &\tabincell{l}{$3\times3,64,1$\\$3\times3,64,2$\\$3\times3,M,2$}&\tabincell{c}{$\left[\begin{array}{cc}3\times3,&64\\3\times3,&64\end{array}\right]\times4$}&\tabincell{c}{$\left[\begin{array}{cc}3\times3,&64\\3\times3,&64\end{array}\right]\times4$}&\tabincell{c}{$\left[\begin{array}{cc}3\times3,&128\\3\times3,&128\end{array}\right]\times8$}&\tabincell{c}{$\left[\begin{array}{cc}3\times3,&256\\3\times3,&256\end{array}\right]\times3$}&\tabincell{l}{12.36M\\51.89}\\ \hline \hline
HRNet18&\tabincell{l}{$3\times3,64,2$\\$3\times3,64,2$}&\tabincell{c}{$\left[\begin{array}{cc}3\times3,&36\\3\times3,&36\end{array}\right]\times4$}&\tabincell{c}{$\left[\begin{array}{cc}3\times3,&64\\3\times3,&64\end{array}\right]\times4$}&\tabincell{c}{$\left[\begin{array}{cc}3\times3,&72\\3\times3,&72\end{array}\right]\times16$}&\tabincell{c}{$\left[\begin{array}{cc}3\times3,&144\\3\times3,&144\end{array}\right]\times12$}&\tabincell{l}{9.56M\\21.71}\\ \hline
\tabincell{l}{BH-\\HRNet18}&\tabincell{l}{$3\times3,64,1$\\$3\times3,64,2$\\$3\times3,M,2$}&\tabincell{c}{$\left[\begin{array}{cc}3\times3,&36\\3\times3,&36\end{array}\right]\times4$}&\tabincell{c}{$\left[\begin{array}{cc}3\times3,&36\\3\times3,&36\end{array}\right]\times4$}&\tabincell{c}{$\left[\begin{array}{cc}3\times3,&72\\3\times3,&72\end{array}\right]\times12$}&\tabincell{c}{$\left[\begin{array}{cc}3\times3,&144\\3\times3,&144\end{array}\right]\times6$}&\tabincell{l}{5.85M\\21.66}\\
\bottomrule[1.5pt]
\end{tabular}}
\tabcolsep5pt
\end{table*}

Optimizing the loss function is an effective strategy to improve the detection performance of tiny objects. Zhu \etal \cite{tinaface} modify the loss function of bounding box regression from the widely applied smooth L1 loss \cite{fastrcnn} to DIoU loss \cite{diou}. It ensures the loss function is consistent with the objective of bounding box regression by including the IoU metrics. 

Our proposed approach, of modifying the backbone, is orthogonal to all this previous work. Hence, potentially all these diverse techniques might be further improved by combining them with our proposed approach.

\section{\uppercase{Methodology}}

% problem and proposals
A natural idea to preserve the high-resolution features required for tiny object detection is to dispense with several down-samplings. However, such a simplistic approach would significantly increase the computational complexity of the backbone as the number of calculations increases quadratically with the size of a feature-map. To avoid this issue, our proposed bottom-heavy (BH) architectures perform as much down-sampling as the corresponding original architecture, but the number of convolutional filters is increased in the earlier stages of the network, and decreased in the later stages. This decrease in the number of convolutional filters in the later layers reduces the amount of computation performed in the deeper layers to compensate for the increased computation performed in the earlier layers. Specifically, in our BH architecture, the number of convolutional layers is increased before the feature-map is down-sampled for the third time, and decreased after this point. The proportion of FLOPs performed in different stages of a ResNet50 and a BH-ResNet50 is shown in Fig~\ref{fig:proportion}. The change in the distribution of the convolutional layers is designed to ensure that the original backbone and the modified version perform the same number of floating-point operations, to allow a fair comparison of the proposed backbones with existing ones. Another result of these modifications is that the BH networks have fewer parameters than their original counter-parts. Full details are provided in Table~\ref{tab:shallow}. These modifications ensure that high-resolution information that can distinguish tiny objects is processed more thoroughly, using more convolutional layers.

\begin{table}[tb]
\centering
\caption{Classification top-1 accuracy of backbone architectures after pre-training on CIFAR-100, ImageNet 32 and ImageNet.}
\label{tab:pretrain}
\resizebox{\columnwidth}{!}{%
\tabcolsep2.5pt
\begin{tabular}{l|r|r|r}
\toprule[1.5pt]
\textbf{Method}&\textbf{CIFAR-100}&\textbf{ImageNet32}&\textbf{ImageNet} \\ \hline
ResNet50&65.82\%&43.12\%&75.92\%\\
BH-ResNet50&68.57\%&48.75\%&74.68\%\\ \hline
HRNet18&55.75\%&---&76.80\%\\ 
BH-HRNet18&59.91\%&50.68\%&76.62\%\\ \hline
HRNet32&55.08\%&---&78.50\%\\ 
BH-HRNet32&61.71\%&56.00\%&78.23\%\\
\bottomrule[1.5pt]
\end{tabular}%
\tabcolsep5pt
}
\end{table}

\section{\uppercase{Experiments}}

\begin{table*}[tb]
\centering
\caption{Detection performance of previous methods and our modified (BH) architecture on TinyPerson. Pre-training is done on CIFAR-100.}
\label{tab:eval_cifar}
\resizebox{\textwidth}{!}{%
\tabcolsep2.5pt
\begin{tabular}{ll|r|r|r|r|r|r}
\toprule[1.5pt]
\textbf{Method}&\textbf{Backbone}&\textbf{$mAP_{tiny}^{50}$}&\textbf{$mAP_{tiny1}^{50}$}&\textbf{$mAP_{tiny2}^{50}$}&\textbf{$mAP_{tiny3}^{50}$}&\textbf{$mAP_{small}^{50}$}&\textbf{FLOPs} \\ \hline
Faster R-CNN-FPN \cite{fasterrcnn}&ResNet50&43.02&26.09&47.43&54.24&\textbf{60.13}&75.58\\ 
&BH-ResNet50&\textbf{46.53}&\textbf{31.65}&\textbf{50.51}&\textbf{56.60}&59.45&75.79\\  \cline{2-8}
&HRNet32&37.67&24.62&41.84&47.29&51.39&100.66\\ 
&BH-HRNet32&\textbf{41.82}&\textbf{27.19}&\textbf{46.03}&\textbf{51.85}&\textbf{55.39}&100.62\\ \cline{2-8}
&HRNet18&36.94&24.05&41.46&45.94&50.57&69.33\\ 
&BH-HRNet18&\textbf{40.27}&\textbf{25.56}&\textbf{44.50}&\textbf{50.18}&\textbf{53.65}&69.28\\ \hline
Faster R-CNN-SM \cite{tinyperson}&ResNet50&43.21&27.12&47.44&54.52&59.78&75.58\\ 
&BH-R50&\textbf{46.69}&\textbf{31.19}&\textbf{50.88}&\textbf{57.13}&\textbf{61.09}&75.59\\ \hline
Adap-FCOS \cite{tov_github}&ResNet50&36.29&21.70&40.60&46.10&49.87&174.74\\
&BH-ResNet50&\textbf{41.42}&\textbf{25.92}&\textbf{45.49}&\textbf{51.33}&\textbf{57.08}&174.95\\ \hline
Faster R-CNN-RFLA \cite{rfla}&ResNet50&42.41&24.73&45.39&55.11&61.50&75.59\\ 
&BH-ResNet50&\textbf{45.34}&\textbf{27.09}&\textbf{49.27}&\textbf{58.41}&\textbf{62.63}&75.58\\ \hline
RetinaNet-SPPNet \cite{sspnet}&ResNet50&48.37&\textbf{36.78}&54.47&55.05&60.74&260.06\\ 
&BH-ResNet50&\textbf{49.23}&36.01&\textbf{54.47}&\textbf{56.17}&\textbf{61.60}&260.27\\ \hline
Cascade R-CNN-SSPNet \cite{sspnet}&ResNet50&48.79&35.59&50.75&\textbf{57.59}&\textbf{65.63}&186.62\\
&BH-ResNet50&\textbf{48.96}&\textbf{37.69}&\textbf{52.15}&55.69&62.62&186.83\\
\hline
Faster R-CNN-SSPNet \cite{sspnet}&ResNet50&48.05&36.40&50.38&56.51&62.94&158.82\\ 
&BH-ResNet50&\textbf{49.78}&\textbf{39.99}&\textbf{52.26}&\textbf{57.14}&\textbf{64.44}&159.03\\ 
\bottomrule[1.5pt]
\end{tabular}%
\tabcolsep5pt
}
\end{table*}

\begin{table*}[t]
\caption{Detection performance of previous methods and our modified (BH) architecture on TinyPerson. Pre-training is done on ImageNet and downsampled ImageNet (indicated with $\dagger$). Only results with our proposed backbone are trained by ourselves while all others are previously reported results.}
\label{tab:eval_in1k}
\resizebox{\textwidth}{!}{%
\tabcolsep2.5pt
\begin{tabular}{ll|r|r|r|r|r|r}
\toprule[1.5pt]
\textbf{Method}&\textbf{Backbone}&\textbf{$mAP_{tiny}^{50}$}&\textbf{$mAP_{tiny1}^{50}$}&\textbf{$mAP_{tiny2}^{50}$}&\textbf{$mAP_{tiny3}^{50}$}&\textbf{$mAP_{small}^{50}$}&\textbf{FLOPs} \\ \hline
Faster R-CNN-FPN \cite{fasterrcnn}&ResNet50$\dagger$&47.81&31.78&53.54&58.34&64.56&75.58\\
&ResNet50&47.35&30.25&51.58&58.95&63.18&75.58\\ 
&BH-ResNet50$\dagger$&\textbf{52.60}&\textbf{38.18}&\textbf{58.18}&\textbf{61.66}&\textbf{67.42}&75.59\\
&BH-ResNet50&52.03&37.95&57.58&60.99&65.71&75.59\\  \cline{2-8}
% &BH-ResNet50&\textbf{49.91}&\textbf{33.47}&\textbf{57.26}&\textbf{59.90}&\textbf{64.76}&75.59\\ \cline{2-8}
&HRNet32&53.05&35.61&58.46&63.87&68.02&100.66\\ 
&BH-HRNet32$\dagger$&52.62&\textbf{38.76}&57.64&62.28&67.42&100.62\\
&BH-HRNet32&\textbf{53.29}&38.08&\textbf{58.92}&\textbf{63.08}&\textbf{67.72}&100.62\\ \cline{2-8}
%&BH-HRNet32&\textbf{54.27}&\textbf{37.27}&\textbf{60.44}&\textbf{64.84}&\textbf{68.44}&100.62\\ \cline{2-8}
&HRNet18&52.28&34.92&58.39&63.36&\textbf{67.86}&69.33\\ 
&BH-HRNet18$\dagger$&50.04&35.43&55.10&60.04&65.03&69.28\\
&BH-HRNet18&\textbf{52.69}&\textbf{36.12}&\textbf{59.13}&\textbf{63.48}&67.45&69.28\\ \hline
Faster R-CNN-SM \cite{tinyperson}&ResNet50$\dagger$&50.65&33.68&55.69&61.40&\textbf{67.39}&75.58\\
&ResNet50&51.33&33.91&55.16&\textbf{62.58}&66.96&75.58\\ 
&BH-ResNet50$\dagger$&\textbf{52.04}&\textbf{37.87}&57.01&61.32&65.91&75.59\\
&BH-ResNet50&51.48&37.35&\textbf{57.35}&60.49&65.47&75.59\\ \hline
Adap FCOS \cite{tov_github}&ResNet50$\dagger$&42.13&22.62&46.50&54.32&57.61&174.74\\
&ResNet50&47.42&29.05&52.06&59.15&65.06&174.74\\ 
&BH-ResNet$\dagger$&48.20&28.56&53.00&59.87&66.51&174.95\\
&BH-ResNet&\textbf{49.16}&\textbf{31.13}&\textbf{53.89}&\textbf{60.09}&\textbf{67.60}&174.95\\ \hline
Faster R-CNN-RFLA \cite{rfla}&ResNet50 $\dagger$&45.78&27.72&49.09&58.33&64.38&75.58\\ 
&ResNet50&48.86&30.35&54.15&61.28&66.69&75.58\\ 
&BH-ResNet50 $\dagger$&50.07&32.29&54.83&62.60&\textbf{68.21}&75.59\\
&BH-ResNet50&\textbf{52.27}&\textbf{34.36}&\textbf{56.62}&\textbf{64.48}&67.77&75.59\\ \hline
RetinaNet-SPPNet \cite{sspnet}&ResNet50$\dagger$&51.28&39.38&56.60&58.55&65.01&260.06\\
&ResNet50&50.98&38.97&57.26&57.68&63.38&260.06\\ 
&BH-ResNet50$\dagger$&\textbf{55.34}&\textbf{42.45}&\textbf{61.04}&\textbf{62.36}&67.63&260.27\\ 
&BH-ResNet50&52.70&39.61&58.82&60.35&\textbf{69.00}&260.27\\ \hline
Cascade R-CNN-SSPNet \cite{sspnet}&ResNet50$\dagger$&54.31&39.25&56.81&63.54&69.40&186.62\\
&ResNet50&56.60&42.41&59.56&65.36&\textbf{71.20}&186.62\\
&BH-ResNet50$\dagger$&56.54&43.65&59.54&64.45&69.59&186.83\\ 
&BH-ResNet50&\textbf{57.89}&\textbf{44.39}&\textbf{60.86}&\textbf{66.25}&71.03&186.83\\
\hline
Faster R-CNN-SSPNet \cite{sspnet}&ResNet50$\dagger$&50.48&36.81&51.08&60.87&67.38&158.82\\
&ResNet50&58.58&44.92&\textbf{62.22}&67.19&71.88&158.82\\
&BH-ResNet50$\dagger$&56.22&46.07&58.03&63.36&69.54&159.03\\
&BH-ResNet50&\textbf{58.97}&\textbf{47.22}&61.61&\textbf{67.45}&\textbf{72.37}&159.03\\
\bottomrule[1.5pt]
\end{tabular}%
\tabcolsep5pt
}
\end{table*}

\subsection{Pre-training}
% why pre-training is necessary
Pre-training is a process that initializes model parameters with previously learned ones. For object detection tasks, it is common to start training the whole model with a backbone that has been pre-trained on an image classification task, typically ImageNet. %In this way, the downstream tasks can gain benefits from previous experience. 
Compared with training the whole detection network from scratch, using a pre-trained backbone is better to maintain a stable and fast training process \cite{pretrain1, pretrain2}. Additionally, the learned knowledge from classification also provides a solid foundation for detection. Our backbones are pre-trained on CIFAR-100 \cite{cifar100}, ImageNet \cite{in1k} and a down-sampled variant of ImageNet, ImageNet32 \cite{dspd_in1k}.

% what is cifar100
The CIFAR-100 dataset \cite{cifar100} consists of 60000 32$\times$32 colour images in 100 classes, separated into 50000 training and 10000 test images.
% why use cifar100
It is chosen for pre-training because: (1) the image size of CIFAR-100 is consistent with the size of target objects in the tiny object detection datasets (2) it has a reasonable training time that is appropriate for prototyping a new backbone; (3) some of the categories in CIFAR-100 such as man, woman, baby, boy, girl, sea etc. are similar to those in the TinyPerson and WiderFace datasets, and hence, it is expected that the parameters will transfer from image recognition to object detection. 

% use imagenet and imagenet32
We also perform experiments using backbones pre-trained on ImageNet. This allows us to compare the performance of the new architecture with published results for existing object detectors that are pre-trained on ImageNet. ImageNet is a large visual dataset containing more than 20000 categories and 14 million images. It is considered a standard choice to pre-train a backbone for object detection tasks. In addition, we also repeat some experiments on a down-sampled variant of ImageNet, named ImageNet32. ImageNet32 contains exactly the same number of images as the original ImageNet, but all images are resized to 32$\times$32 pixels. This dataset has the same advantages as CIFAR100, but provides more training images.

For data augmentation, we use random horizontal flipping, random cropping and resizing.
% training hyper-parameters
The classification results at the end of pre-training are shown in Table~\ref{tab:pretrain}. It can be seen that the proposed BH networks produce more accurate classifications for the datasets with small image sizes, compared to the corresponding original networks. This supports our claim that these bottom-heavy networks are more appropriate for the discrimination of small objects. This improvement in performance on small objects comes at the cost of a small decrease in classification accuracy for large images.
There may be scope for improving these pre-training results (and as a consequence improving the image detection results) as no great effort has been taken to optimise the hyper-settings used during pre-training: we had insufficient computational resources to do so, and searching for the best settings is not the main point of this paper.

\subsection{Fine-tuning}
We trained object detectors using the TinyPerson and WiderFace datasets. TinyPerson has 1610 large-scale seaside images taken from a long distance. It contains 794 and 816 images for training and testing, respectively. The average absolute size of target objects is about $18$ pixels and non-small objects (absolute size$ >32$ pixels) are rare. TinyPerson divides objects into different intervals based on the absolute size (measured in pixels): tiny $[2, 20]$, small $[20, 32]$ and all $[2, \infty]$. The tiny set is partitioned into three (overlapping) subsets: tiny1 $[2, 8]$, tiny2 $[8, 12]$ and tiny3 $[12, 20]$. The WiderFace dataset is a face detection benchmark containing 32,203 images of different human faces with a high degree of variability in scale, pose and occlusion. The average size of objects in WiderFace is about $32$ pixels. It separates objects into three evaluation sets (easy, medium and hard) based on the difficulty of detecting them.

The resolution of the images in TinyPerson ranges from 497 $\times$ 700 pixels to 4064 $\times$ 6354 pixels. Directly feeding large images into the backbone would produce out-of-memory issues. Meanwhile, simply resizing such images to make them more manageable would risk losing information about tiny objects. Therefore we follow the procedure described in \cite{tinyperson} for training with this dataset. Specifically, images are cut into overlapping patches of size 640 pixels$\times$512 pixels with 30 pixels overlap. We use the default training schedule for Faster-RCNN and Adap-FCOS\cite{tov_github} on TinyPerson: 12 epochs with stochastic gradient descent (SGD) as the optimizer. The learning rate is initialized to 0.01 and decreased by a factor of 0.1 after 8 and 11 epochs for training on 2 GPUs. The only difference for training different architectures of SSPNet is that the default schedule is 10 epochs.
% settings of WiderFace
For WiderFace, the input size is 640 pixels$\times$640 pixels. The default training schedule of 630 epochs with SGD \cite{tinaface} is used. The learning rate is initialized to 0.00375 and follows a cosine restart scheduler that restarts the learning rate every 30 epochs. % for training on 3 GPUs.
We do not apply any complicated image augmentation methods but follow the same settings used by the corresponding state-of-the-art methods. 

\section{\uppercase{Results}}

\subsection{Experimental Setup and Evaluation Metrics}

Mean average precision (mAP) is a standard criterion for evaluation in object detection. AP calculates the average precision among a set of different recalls. mAP is the mean AP among all classes. Whether a bounding box is marked as a positive or negative prediction is determined by whether the intersection-over-union (IoU) of the predicted and ground-truth bounding boxes is greater than a set threshold. The $mAP$ on a set of objects $i$ with the percentage IoU threshold $j$ is noted as $mAP_{i}^{j}$. For example, $mAP_{tiny}^{50}$ represents the mAP among objects in the $tiny$ interval (absolute size $\in [2, 20]$) with IoU threshold $0.5$. Floating-point operations (FLOPs) are used to measure the computational complexity of a model. The higher the number of FLOPs, the more calculations are performed by the method, for a given size of the image. In Tables~\ref{tab:eval_cifar} and \ref{tab:eval_in1k}, FLOPs are measured for an image size of 640$\times$512 pixels.

\begin{table*}[t]
\centering
\caption{The difference of $mAP_{tiny}$ between the standard backbones and our 'Bottom Heavy' versions backbones, calculated by calculating $mAP_{tiny}$ (for the BH architecture) minus $mAP_{tiny}$ (for the corresponding standard architecture).}
\label{tab:increase}
%\resizebox{\columnwidth}{!}{%
\begin{tabular}{l|r|r|r}
\toprule[1.5pt]
\textbf{Method}&\textbf{CIFAR-100}&\textbf{ImageNet32}&\textbf{ImageNet} \\ \hline
Faster R-CNN \cite{fasterrcnn}&+3.51&+4.79&+4.68\\
Faster R-CNN-SM \cite{tinyperson}&+3.48&+1.39&+0.15\\ 
Adap FCOS \cite{tov_github}&+5.13&+6.07&+1.74\\ 
Faster R-CNN-RFLA \cite{rfla}&+2.93&+4.29&+3.41\\
RetinaNet-SPPNet \cite{sspnet}&+0.86&+4.06&+1.72\\
Cascade R-CNN-SSPNet \cite{sspnet}&+0.17&+2.23&+1.29\\ 
Faster R-CNN-SSPNet \cite{fasterrcnn}&+1.73&+5.74&+0.39\\ \hline
Average change in $mAP_{tiny}$ &+2.54&+3.96&+2.04\\
\bottomrule[1.5pt]
\end{tabular}%
%}
\end{table*}

\begin{table*}[t]
\centering
\caption{Detection performance of previous methods and our modified architecture on WiderFace. Pre-training is done on ImageNet. BN, GN \& DCN indicates the batch normalization, group normalization and deformable convolution networks respectively.}
\label{tab:eval_widerface_in1k}
%\resizebox{\textwidth}{!}{%
\begin{tabular}{ll|r|r|r|r}
\toprule[1.5pt]
\textbf{Method}&\textbf{Backbone}&\textbf{$mAP_{easy}$}&\textbf{$mAP_{medium}$}&\textbf{$mAP_{hard}$}&\textbf{FLOPs} \\ \hline
TinaFace w BN\cite{tinaface}&ResNet50&\textbf{95.77}&95.43&92.23&191.30\\
&BH-ResNet50&95.56&\textbf{95.71}&\textbf{92.28}&191.54\\ \hline
TinaFace w GN \& DCN \cite{tinaface}&ResNet50&96.27&95.67&93.06&184.29\\
&BH-ResNet50&\textbf{96.61}&\textbf{95.98}&\textbf{93.26}&188.60\\
\bottomrule[1.5pt]
\end{tabular}%
%}
\end{table*}

\subsection{Comparison to State-of-the-arts}
To evaluate our proposed backbone, we tested several architectures that have previously been shown to produce state-of-the-art performance on TinyPerson and WiderFace, replacing the standard backbone with our proposed bottom-heavy (BH) equivalent.  
The results for TinyPerson are shown in Tables~\ref{tab:eval_cifar} and \ref{tab:eval_in1k}. In each case better performance was achieved using the proposed backbones than with the original ones, as summarized in Table~\ref{tab:increase}. This was achieved without increasing the computational complexity and using backbones containing fewer parameters than the originals.
% mention the hyperparameter is not ideal for imagenet pretrained
It can also be seen from Table~\ref{tab:increase} that the proposed BH modifications produce a significant improvement in detection performance compared to standard backbones, and that this improvement is highest when using CIFAR-100 and ImageNet32 for pre-training. 

A pre-training dataset should be consistent with the target task \cite{pretrain1}. Down-sampled ImageNet is a simple alternative to ImageNet that has higher consistency with tiny object detection from the perspective of object size. However, from our experiments, only for some methods did pre-training on ImageNet32 outperform pre-training on ImageNet. It is likely that ImageNet32 is more relevant to tiny object detection but the blurring of images caused by downsampling also makes it harder to learn useful representations from this dataset. The low top-1 classification accuracy, as shown in Table~\ref{tab:pretrain}, provides evidence for this. CIFAR-100 also uses images that are consistent in size to the target objects in tiny object detection, but detection is much worse overall when using CIFAR-100 for pre-training, presumably because the small size of this dataset means that the representations that are learnt  generalise more poorly.

When pre-training on ImageNet, although our method still outperforms the state-of-the-art methods, the improvement is not as significant as when pre-training with CIFAR-100 and ImageNet32. This difference is likely due to two reasons. One is that our backbones work better when they are pre-trained on a dataset that is more consistent with the tiny objects because our backbones are designed specifically for such objects. Another reason is we have simply used the same hyper-parameters to pre-train our BH networks on ImageNet as are used for the corresponding standard architectures, as we did not have the computational resources to search for more optimum  hyper-parameters appropriate for pre-training BH networks on such a large dataset. 

The results for WiderFace are shown in Table~\ref{tab:eval_widerface_in1k}. We evaluate performance using the current best method, TinaFace \cite{tinaface}, with two different settings. Our backbones outperform the standard backbones in both cases in medium and hard tasks and have achieved the state-of-the-art result in the easy task at 96.61\%.

Overall, improved performance was generated using different backbone families (ResNet and HRNet), different network depths, pre-training with different datasets (CIFAR-100, ImageNet and down-sampled ImageNet), and when integrating the proposed backbones into several state-of-the-art tiny object detection frameworks. These results demonstrate that the proposed method of network modification generalises, and could potentially be used with other architectures to make them more suitable for small and tiny object classification and detection. Furthermore, our results demonstrate that the use of standard backbones is harmful for tiny object detection. The results also suggest that current architectures are potentially wasting computational resources by performing unnecessary computations in deep layers.

%------------------------------------------------------------------------
\section{\uppercase{Conclusion}}
This paper proposes that standard deep neural network architectures used as feature-ext\-rac\-tion front-ends in object-detection algorithms produce poor features for tiny object detection tasks. Our claim is that early down-sampling in such architectures results in information loss and features that are poor at representing small objects. To test this claim we designed bottom-heavy versions of popular backbone architectures, ResNet and HRNet, that increase the number of convolutional layers in the shallow, high-resolution, stages of the network, and have fewer convolutional layers at later stages in the network. These changes in the distribution of the convolutional layers are made to ensure that the computational complexity of the modified backbones matches that of the original, standard, network. 
Experimental results show that these changes, despite reducing the number of parameters in the networks, result in more accurate object detection across a number of object-detectors, using different backbones, and pre-training schemes (CIFAR100, ImageNet and ImgaeNet32), for two standard benchmark datasets (TinyPerson and WiderFace).
% Our best result get {}\% better than the current state-of-the-art method.
The architectures proposed in this paper are not a final answer to the information loss problem. Rather they serve as a motivation for developing improved feature-extraction backbones appropriate to the task. Hopefully, our current results and insights will inspire the development of even better backbones for tiny object detectors in future research.

%------------------------------------------------------------------------
\section*{\uppercase{Acknowledgements}}

The authors acknowledge the use of the research computing facilities at King’s College London, Rosalind, the King's Computational Research, Engineering and Technology Environment (CREATE) \cite{create}, and the Joint Academic Data science Endeavour (JADE) facility. This research was funded by the King's - China Scholarship Council (K-CSC).

\bibliographystyle{apalike}
{\small
\bibliography{example}}

% \section*{\uppercase{Appendix}}

% If any, the appendix should appear directly after the
% references without numbering, and not on a new page. To do so please use the following command:
% \textit{$\backslash$section*\{APPENDIX\}}

\end{document}